\documentclass[journal]{IEEEtran}

\usepackage{lipsum}
\usepackage[edges]{forest}
\usepackage[utf8]{inputenc} 
\usepackage[T1]{fontenc}    
\usepackage{hyperref}       
\usepackage{url}            
\usepackage{booktabs}       
\usepackage{amsfonts}       
\usepackage{nicefrac}       
\usepackage{microtype}      
\usepackage{lipsum}		
\usepackage{enumerate}
\usepackage{graphicx}
\usepackage{doi}
\usepackage{mathptmx}
\usepackage{amsmath}
\usepackage{color}
\usepackage{tikz}
\usepackage{forest}
\usepackage{enumitem}
\usepackage{color,soul}
\usepackage{amssymb}
\usepackage{svg}
\usepackage{pdfpages}
\usetikzlibrary{trees,positioning,shapes,shadows,arrows}
\usepackage[normalem]{ulem}
\usepackage{multirow, makecell}

\graphicspath{ {./images/} }
\newcommand{\comment}[1]{}

\ifCLASSINFOpdf
\else
\fi
\begin{document}
%
\title{Towards Fairness-Aware Federated Learning}
%
%
%

\author{Yuxin Shi,
        Han Yu\textsuperscript{\rm *},~\IEEEmembership{Senior Member,~IEEE},
        Cyril Leung,~\IEEEmembership{Senior Member,~IEEE}
\thanks{Yuxin Shi is with the School of Computer Science and Engineering,
Nanyang Technological University (NTU), Singapore; Alibaba-NTU Singapore Joint
Research Institute, NTU, Singapore; and Alibaba Group, Hangzhou, China.}
\thanks{Han Yu is with the School of Computer Science and Engineering, Nanyang Technological University (NTU), Singapore.}
\thanks{Cyril Leung is with the Department of Electrical and Computer Engineering, The University of British Columbia, Vancouver, BC, Canada; and Alibaba-NTU Singapore Joint Research Institute, NTU, Singapore.}
\thanks{$^*$Corresponding author: han.yu@ntu.edu.sg}
}

\maketitle

\begin{abstract}
Recent advances in Federated Learning (FL) have brought large-scale collaborative machine learning opportunities for massively distributed clients with performance and data privacy guarantees. However, most current works focus on the interest of the central controller in FL,
and overlook the interests of the FL clients.
This may result in unfair treatment of clients that discourages them from actively participating in the learning process and damages the sustainability of the FL ecosystem. 
Therefore, the topic of ensuring fairness in FL is attracting a great deal of research interest. In recent years, diverse Fairness-Aware FL (FAFL) approaches have been proposed in an effort to achieve fairness in FL from different perspectives. However, there is no comprehensive survey that helps readers gain insight into this interdisciplinary field.
%
This paper aims to provide such a survey. 
By examining the fundamental and simplifying assumptions, as well as the notions of fairness adopted by existing literature in this field, we propose a taxonomy of FAFL approaches covering major steps in FL, including client selection, optimization, contribution evaluation and incentive distribution. In addition, we discuss the main metrics for experimentally evaluating the performance of FAFL approaches, and suggest promising future research directions towards FAFL.
\end{abstract}


%
\IEEEpeerreviewmaketitle

\section{Introduction}

\begin{figure}[ht]
    \centering
    \includegraphics[width=1\linewidth]{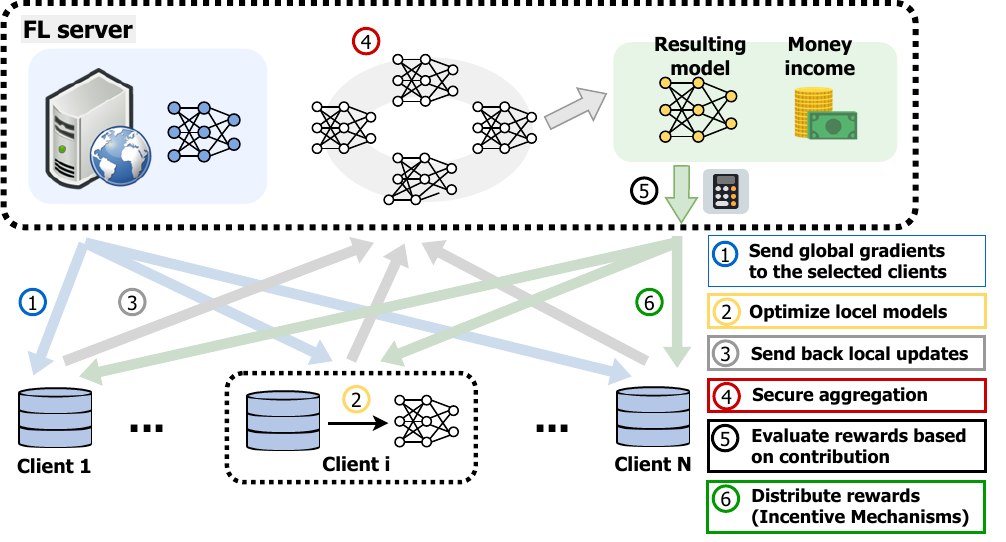}
    \caption{General FL training process involving N clients.}
    \label{fig:FL_process}
    \vspace{-12pt}
\end{figure}

\IEEEPARstart{F}{ederated} learning (FL) 
\cite{yang2019federated,FL:2019} is a new machine learning paradigm for collaboratively training models involving multiple data owners (a.k.a. clients) with the aim to protect data privacy \cite{Anonymous2013ConsumerDP}, shown in Figure \ref{fig:FL_process}. 
Its potential to help the field of artificial intelligence (AI) thrive in privacy-respecting societies has attracted increasing attention from academia and industry alike. During federated model training, the data owners contribute not only their local data, but also computation and communication resources to facilitate collaborative model training. 
As the quality and quantity of the data, as well as the local resources vary among data owners, their contributions to the final FL model will vary. 
This, in turn, may affect the benefits they receive from the data federations they join, in cases where FL incentive mechanisms are deployed \cite{zhan9369019}. 
Ensuring fairness in FL is an emerging research challenge that requires an interdisciplinary approach \cite{kairouz2021advances}. Since the FL server is the controller, it should ideally not make decisions that unfairly favour any particular group of data owners. 
However, in most current FL paradigms, fair treatment of data owners is not given a high priority. Problems of fairness \cite{Zhang-et-al:2022IJCS} can arise throughout the FL training process, as described below.

Firstly, unfairness may arise in the stage of \textit{client selection}. 
Methods have been proposed for client selection in FL, but most of them focus only on the server's interest (e.g., increasing convergence speed \cite{Wang2020OptimizingFL, Nishio2019ClientSF, goetz2019active, ribero2020communicationefficient} or enhancing the model performance \cite{Yoshida2020HybridFLFW}). These methods favor clients who can respond quickly or contribute to better final model performance  \cite{Zeng2021efficientRM}. As a result, clients with weaker capabilities might be excluded from the FL process. This reduces weaker clients' chances of (i) receiving a model that fits their local data distributions well, and (ii) receiving any incentives which might be involved. 
From the server's perspective, the final model may not generalize well because the excluded clients may contain samples that the current model does not cover. 

Secondly, fairness issues can also occur during FL model \textit{optimization}.
When performing the global optimization by aggregating clients' model updates, the server aims to minimize a loss function under a specific distribution \cite{Mohri2019AgnosticFL,Li2020FairRA}. However, this distribution may not be able to capture the diversity of the entire distributed training set \cite{Mohri2019AgnosticFL}, leading to prediction biases on individual clients' local datasets. 

Thirdly, unfairness can occur during FL \textit{incentive distribution}. In FL, rewards (monetary or otherwise) may be needed to compensate for clients' local resources (e.g., computation power, battery power, communication bandwidth) and motivate participation.
Under the standard FL setting, the FL system distributes the same aggregated model as rewards to all the clients.
However, such a scheme ignores the fact that local updates of clients are of inconsistent qualities due to diverse local dataset qualities and training capabilities \cite{Zhang2020EnablingEA}. The scheme can be perceived as unfair by clients who contribute more significantly to the final FL model performance. 
Such a problem, referred to as the free-rider issue in FL \cite{Lyu2019TowardsFA}, can discourage future participation of high-quality FL clients \cite{Zhang2020HierarchicallyFF}, which in turn harms the FL system.

Last but not least, \textit{contribution evaluation} is an essential step in FL to promote fairness in the \textit{client selection} and \textit{incentive distribution} stages. Contribution evaluation in FL aims to estimate the contribution made by each client without exposing its private local data \cite{Zeng2021ACS}. The evaluation can be based on self-reported information such as data quantity \cite{Zhang2020HierarchicallyFF}, data quality \cite{Zhang2020HierarchicallyFF,kang2019incentivecontract}, or observations on marginal model improvement \cite{Song9006327,WangDataValue2020,WeiDataValuation2020}. FL schemes that use client contribution as a reference to select high-quality clients in order to improve model performance are emerging \cite{goetz2019active}. FL incentive schemes also often take client contribution evaluation results into account \cite{Zhang2020HierarchicallyFF,kang2019incentivecontract}. Thus, unfairness in FL client contribution evaluation can have a far-reaching impact throughout the FL pipeline.

These fairness-related issues can have an adverse impact on both the FL clients and the FL server if not handled properly. On the one hand,  unfair treatment can discourage clients from joining FL training. On the other hand, blindly treating clients equally without regard to their potential contributions can reduce the server's ability to attract high-quality clients, resulting in FL models that may not generalize well. Hence, ensuring fairness in FL is essential as it is the key to sustainable healthy collaboration in such an ecosystem.

As FL research is attracting increasingly significant attention, many survey papers have been published. They provide different perspectives on various topics about FL. For example,  \cite{yang2019federated, Li2020survey} provide an overview of general methods and applications of FL. Lim et al. 2020 \cite{Lim2020FederatedLI} reviews the methods and applications of FL in the context of mobile edge networks. In \cite{lyu2020threats, MOTHUKURI2021619}, topics of security and privacy in FL are surveyed. Tan et al. 2022 \cite{tan2021personalized} provide a review of FL personalization techniques for dealing with data heterogeneity. The topic of FL incentive mechanism design is reviewed in \cite{zhan9369019, Zeng2021ACS}. In recent years, diverse Fairness-Aware Federated Learning (FAFL) approaches have been proposed to achieve fairness from different angles. There is currently no comprehensive survey on this topic.

In this paper, we bridge this gap by providing a comprehensive review of the existing literature on FAFL. This paper contributes to the AI literature in the following ways:
\begin{itemize}[leftmargin=*]
    \item We analyze the fundamental and simplifying assumptions commonly adopted in existing FL work and discuss their potential impact on incorporating fairness into FL.
    \item We summarize the main notions of fairness adopted in existing FAFL approaches to provide an overview of the diverse motivations in this field.
    \item We propose a taxonomy of FAFL based on the major steps in FL, and summarize the challenges in each step. To the best of our knowledge, it is the first such taxonomy on this topic, and provides new perspectives to existing works in this field.
    \item We discuss the common evaluation metrics adopted in existing FAFL approaches, thereby, providing readers with a useful guide on experiment design.
    \item We outline promising future research directions towards FAFL approaches. For each direction, we analyze the limitations in the current literature and propose potential ways forward.
\end{itemize}

\section{Common Assumptions}  \label{ComA}
In order to understand the background of existing FAFL approaches, it is important to examine the assumptions made in existing FL approaches. As horizontal federated learning (HFL) \cite{yang2019federated} is currently the most studied FL approach, we focus our discussions on HFL settings.
The main aim of HFL \cite{pmlr-v54-mcmahan17a} is to allow multiple data owners to collaboratively train a shared model while preserving the privacy of sensitive local data. At the beginning of each training round, the FL server first selects a group of data owners as its FL clients, and distributes the latest global model to them. Then, the clients train models locally and upload them to the server for aggregation. 

Existing FL approaches use many assumptions, which can be summarized into two categories:
\begin{enumerate} [leftmargin=*,label=\Alph*.]
    \item \textbf{Fundamental Assumptions:} These assumptions are essential in the operation of FL, and are widely adopted. The most common ones include \cite{kairouz2021advances}:
    \begin{enumerate}[label=\arabic*)]
        \item A client's local data are assumed to be sensitive and shall not be exposed to anyone else.
        \item The server and the clients are self-interested and rational.
    \end{enumerate}
    \item \textbf{Simplifying Assumptions:} These assumptions are made to enable specific FL approaches to operate. They are usually limited to specific scenarios and are not necessarily widely adopted in this field. They include:
    \begin{enumerate}[label=\arabic*)]
        \item \textit{FL clients are trustworthy.} clients are assumed to be honest in that they use their real private data to perform local training and submit the local models truthfully to the FL server. This assumption enables some FL frameworks to trust clients' self-reports about their local data quantity/quality to the FL server for making decisions related to client selection and incentive distribution \cite{Lyu2019TowardsFA,Zhang2020HierarchicallyFF,motivatingstackberg, Zeng2020FMoreAI}. 
        \item \textit{The FL server is trustworthy.} The server often plays a central role in FL. It is assumed that the server can be trusted to make reliable decisions when selecting clients and distributing incentives \cite{Zhang2020HierarchicallyFF,motivatingstackberg,Zeng2020FMoreAI}.
        \item \textit{Data owners always agree to join FL.} When the server sends out invitations to data owners to join FL, it is assumed that they will accept the invitations \cite{Nishio2019ClientSF}.
        \item \textit{Prior information about the clients is available.} Some works assume that the prior information (e.g., client capabilities, data quality, past task performance, etc.) can be tracked by the FL server \cite{Yu2020AFI,Ye2020FederatedLI}, or obtained by the FL server from a third party (e.g., a blockchain) \cite{Lyu2019TowardsFA,Rehman2021TrustFedAF}. This information is then utilized for subsequent client selection or incentive distribution.
        \item \textit{A client dedicates all its resources to a given FL training task.} After a client joins FL training, it is assumed that it will dedicate all its computational resources to the learning task (i.e., a client only joins one data federation at any given time).
        \item \textit{Clients' local data remains unchanged during FL training.} An FL task typically requires multiple training rounds (i.e., the system repeats local model training and global model aggregation until a desired training accuracy is achieved). It is commonly assumed that clients' local data do not change during FL training \cite{Wang2020OptimizingFL}.
        \item \textit{Monopoly FL server.} Most existing FL works implicitly assume that there is only one data federation (i.e., a monopoly) in a given application scenario \cite{kang2019incentivecontract}.
    \end{enumerate}
\end{enumerate}

While the fundamental assumptions remain unchanged, some of the simplifying assumptions have been relaxed to include FAFL approaches for achieving various notions of fairness.


\section{Notions of Fairness in FAFL} 
\label{fairnessnotions}
Fairness has been studied in many disciplines. In machine learning, dozens of fairness notions have been proposed \cite{Mehrabi-et-al:2021}. Each notion focuses on a particular aspect and the interest of a specific group of stakeholders. Thus, it is not suitable to compare the relative merits of these fairness notions. In this section, we provide an overview of fairness notions adopted by existing FAFL approaches, focusing on their motivations and the stakeholders in the FL paradigm that they serve. 

\begin{enumerate} [leftmargin=*]
    \item \textbf{Performance Distribution Fairness:} This notion of fairness aims to generalize standard accuracy parity\cite{Mehrabi-et-al:2021} by measuring the degree of uniformity in performance across FL client devices \cite{Li2020FairRA}. 
    Model $w$ is more fair than model $\hat{w}$ if the performance of model $w$ on the $n$ devices is more uniform than that of model $\hat{w}$.
    The mean and the variance of the accuracy values are used to measure the degree of uniformity. It is an individual-level fairness.
    
    \item \textbf{Good-Intent Fairness}: This notion of fairness minimizes the maximum loss for the underlying protected groups so as to avoid overfitting any particular model at the expense of others \cite{Mohri2019AgnosticFL}. This notion is applied to the scenario where the dataset is split into different groups. This notion optimizes the group with the worst performance, hence reducing the variance of the accuracy values across all the groups. 
    
    \item \textbf{Group Fairness}: This notion of fairness aims to minimize the disparities in algorithmic decisions across different groups \cite{Du2020FairnessawareAF}. The disparity in algorithmic decisions can be measured by \textit{demographic parity} \cite{demographic2012} and \textit{equal opportunity} fairness notions \cite{EqualOpp2016}.

    \item \textbf{Selection Fairness}: This notion of fairness aims to mitigate the bias in an FL model by increasing the chance of participation for under-represented or never-represented clients \cite{Zhou2021LossTF}. One existing approach to ensure this is through setting the sampling constraints (e.g., long-term fairness constraint \cite{Huang2021AnEC}). Another possible approach is to set the clients' selection probabilities proportional to their cost and their potential contribution to the FL model. It is an individual-level fairness for client selection.
    \item \textbf{Contribution Fairness}: Under this notion of fairness which has its foundation in Game Theory \cite{Rabin1993IncorporatingFI}, a client's payoff shall be proportional to its contribution to the FL model. It is a type of distribution fairness that is not concerned with optimizing the FL model accuracy. Rather, it is an individual-level fairness, often adopted by FL incentive schemes to guide reward allocation \cite{gametheoretic2020,LyuCollaborative2020}.
    \item \textbf{Regret Distribution Fairness}: This notion of fairness aims to minimize the difference of the regret among FL clients as a result of waiting to receive incentive payout \cite{Yu2020AFI}. Regret refers to the difference between what the data owner has received so far and what he is supposed to receive while taking into account how long he has been waiting to receive the full payoff. This notion is an individual-level fairness.
    \item \textbf{Expectation Fairness}: This notion of fairness builds on top of the regret distribution fairness, which is also an individual-level fairness. It aims to minimize the inequity among the clients at different points in time as incentive rewards are gradually paid out over a period of time \cite{Yu2020AFI}. Regret distribution fairness and expectation fairness are useful in situations where the incentive budget is derived from future earnings of the FL model, and the participating FL clients are gradually compensated as earnings are generated.
\end{enumerate}


Good-Intent Fairness and Group Fairness are designed to protect the interest of the eventual users of an FL model. Selection Fairness and Performance Distribution Fairness are primarily designed to serve the interest of the FL clients while aiming to enhance FL model performance during the process (thereby also protecting the interest of the FL server and the eventual users of the FL model). Contribution Fairness, Regret Distribution Fairness and Expectation Fairness are designed to guide FL incentive mechanisms to take fairness into account. Contribution Fairness is commonly adopted by both monetary and non-monetary FL incentive schemes \cite{Zeng2021ACS}, while Regret Distribution Fairness and Expectation Fairness are used for situations in which incentives are derived from future earnings.

Existing notions of fairness in FL can also be categorized
from two perspectives, as shown in Table \ref{table:fairness_categories}: 1) \textit{Performance fairness} and \textit{Cooperation fairness}, which are categorized regarding different training steps; 2) \textit{Individual-level fairness} and \textit{Group-level fairness}, which are categorized regarding different target entities.


\begingroup

\setlength{\tabcolsep}{5pt} 
\renewcommand{\arraystretch}{1.4} 
\begin{table}[]
\centering
\caption{Categories of fairness notions in FAFL research.}
\resizebox*{0.95\linewidth}{!}{
\begin{tabular}{l|ccc}
\hline
\multicolumn{1}{c|}{\multirow{2}{*}{\textbf{Notion}}} & \multicolumn{3}{c}{\textbf{Category}}                                                                                                                         \\ \cline{2-4} 
\multicolumn{1}{c|}{}                                 & \multicolumn{2}{c|}{\textbf{Training Stage}}                                                                              & \textbf{Target Entity}          \\ \hline
Good-Intent Fairness                                  & \multicolumn{1}{c|}{\multirow{3}{*}{\makecell{Model\\optimization}}}     & \multicolumn{1}{c|}{\multirow{3}{*}{\makecell{Performance\\fairness}}} & \multirow{2}{*}{Group-level}      \\ \cline{1-1}
Group Fairness                                        & \multicolumn{1}{c|}{}                                        & \multicolumn{1}{c|}{}                                      &                                   \\ \cline{1-1} \cline{4-4} 
\makecell[tl]{Performance Distribution\\Fairness}                     & \multicolumn{1}{c|}{}                                        & \multicolumn{1}{c|}{}                                      & \multirow{5}{*}{Individual-level} \\ \cline{1-3}
Selection Fairness                                    & \multicolumn{1}{c|}{Client selection}                        & \multicolumn{1}{c|}{\multirow{4}{*}{\makecell{Cooperation\\fairness}}} &                                   \\ \cline{1-2}
Contribution Fairness                                 & \multicolumn{1}{c|}{\multirow{3}{*}{\makecell{Incentive\\distribution}}} & \multicolumn{1}{c|}{}                                      &                                   \\ \cline{1-1}
\makecell[tl]{Regret Distribution\\Fairness}                          & \multicolumn{1}{c|}{}                                        & \multicolumn{1}{c|}{}                                      &                                   \\ \cline{1-1}
Expectation Fairness                                  & \multicolumn{1}{c|}{}                                        & \multicolumn{1}{c|}{}                                      &                                   \\ \hline
\end{tabular}
}
\label{table:fairness_categories}
\end{table}
\endgroup

For the first perspective, performance fairness aims to encourage similar performance across different clients or subgroups formed based on sensitive attributes. Performance fairness usually occurs in the \textit{Model Optimization} stage of FL.
Cooperation fairness is fairness in the interaction between server and clients, which aims to promote fair treatment of FL clients and attract more clients to participate in training. This approach of fairness usually occurs in \textit{Client Selection} and \textit{Incentive Distribution} stages.

For the second perspective, individual-level fairness is proposed to encourage the model to have similar performance over different clients. In contrast, group fairness is designed to eliminate prejudice towards a specific group of stakeholders. Different from centralized training, groups in FL can be formed not only based on their sensitive attributes (e.g., gender, race) \cite{Du2020FairnessawareAF} but also on other variables (e.g., class label). Moreover, the groups can also be newly added clients. Group-level fairness can also be degenerated into individual-level fairness by treating every device as a group.


\section{The Proposed Taxonomy of FAFL}
\begin{figure}[ht]
    \centering
    \includegraphics[width=0.95\linewidth]{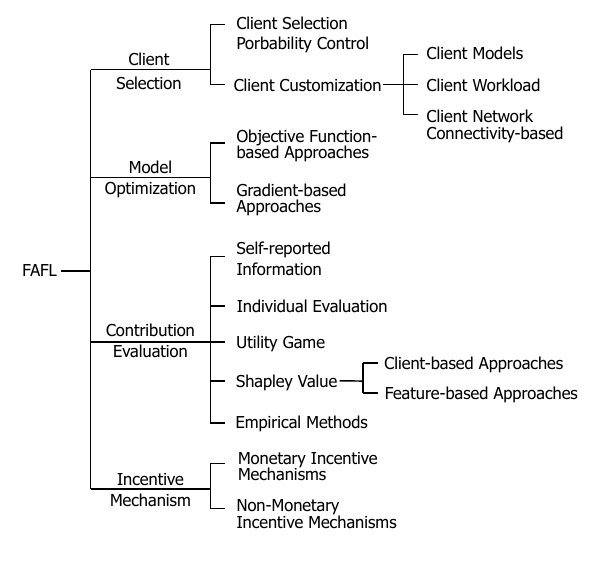}
    \caption{The proposed taxonomy of FAFL approaches.}
    \label{fig:taxonomy1}
\end{figure}

Based on the discussion of the common assumptions and notions of fairness adopted by FAFL approaches, we propose the taxonomy shown in Figure \ref{fig:taxonomy1}.
Existing studies are classified according to the major aspects of FL, including client selection, the model optimization process, contribution evaluation, and incentive mechanism. Some FAFL approaches address several aspects simultaneously, while others focus on a single aspect. The studies in contribution evaluation are also related to the problem of client selection and incentive mechanism, as the client contribution is often used as a basis to invite data owners and distribute the rewards to each client. In this section, we discuss existing works following this taxonomy, highlighting the fairness notions adopted, their approaches and limitations.


\subsection{Fairness in FL Client Selection} \label{clientselection}

One source of potential unfair treatment of FL clients is the FL client selection approach. Many approaches focus on serving the FL server's interest (e.g., increasing convergence speed \cite{Wang2020OptimizingFL} or enhancing the model accuracy \cite{Yoshida2020HybridFLFW}), while ignoring the interests of the FL clients. These works typically use threshold-based approaches to select FL clients (e.g., \textit{FedCS} \cite{Nishio2019ClientSF}) Such approaches commonly use thresholds (e.g., transmission speed, bandwidth, local accuracy, etc.) to filter out unqualified clients and select high-quality clients. 
For example, mobile edge FL systems are often sensitive to network transmission rates. In such cases, data owners with higher transmission rates are more likely to be selected, while those with consistently poor channel conditions may never be selected. However, this brings up another problem of oversampling of clients from specific groups, which leads to the global FL model biasing towards the data
owned by these clients, hence causing model performance deterioration \cite{Cho2020ClientSI}. Moreover, these approaches threaten the interests of clients. The continued participation of clients is the key to maintaining the long-term sustainable operation of an FL system. However, under the threshold-based approaches, clients with the relatively poor quality might not get the chance to join FL (a.k.a. unfair selection among clients). As a result, they cannot receive any rewards. This drives these clients to leave the system, resulting in the attrition of FL participants. Hence, it is crucial to pursue fairness in FL client selection.

It is important to note that fair client selection does not mean selecting everyone with equal probabilities. The heterogeneity among clients also needs to be considered. To achieve fairness in client selection, FAFL approaches need to strike a balance between the interests of the FL server and those of the FL clients. In this section, we review existing FAFL client selection approaches. These works can be divided into two categories: 1) considering fairness factors to ensure that each client has a reasonable probability to be selected, and 2) customizing the model and training procedure for each client.

\subsubsection{FL Client Selection Probability Control}
To mitigate the bias against FL clients with lower computational capabilities or smaller datasets, recent works study the design of sampling constraints for addressing fair client selection. They consider fairness factors to allow less-frequently selected clients to join FL training more often, hence reducing selection bias.

In \cite{Huang2021AnEC}, the authors introduced a long-term fairness constraint towards achieving fair FL client selection. This constraint applies a constant fairness parameter to ensure that the average participation rate of every client is no less than the expected guaranteed rate. A Lyapunov optimization-based framework has been proposed to transfer the original offline problem into an online optimization problem, where FL clients' participation rates are optimized through a queuing dynamics approach. 
Similar to \cite{Huang2021AnEC}, \cite{HuangCS2022} also utilizes a fairness constraint to reserve a certain probability of selection for each client. However, instead of using dynamic queues, \cite{HuangCS2022} adapts the Exp3 algorithms \cite{2002exp3} for the adversarial bandit to calculate the selection probability for each round. Moreover, in \cite{HuangCS2022}, the fairness parameter that is used to determine the selection probability in each round is not necessarily the same.
The results of both works show that a fairer strategy for FL client selection can improve the final accuracy at the cost of training efficiency due to more constraints being involved.


Yang et al. 2020 \cite{Yang2020FederatedLW} also aimed to promote the selection of less-frequently selected clients. They formulated the client selection problem as a Combinatorial Multi-Armed Bandit (CMAB) problem. Each arm represents one client. Its reward is defined by the class distribution of its raw data. The reward of a super arm depends on the rewards of all the pulled arms. While frequently selected clients are regarded as more trustworthy and receive a higher payoff, the system also provides opportunities for less-frequently engaged clients to join FL by increasing their estimated rewards.


One shortcoming of \cite{Huang2021AnEC, HuangCS2022, Yang2020FederatedLW} is that they do not take into account the real-time contribution of individual clients when designing the fairness factors. This shortcoming has been addressed by \cite{SongReputation22}. They proposed a reputation-based client selection policy with fairness constraints. A reputation table is maintained for all clients based on their historical behaviours. A fairness parameter is used to control the trade-off between reputation and the number of successful transmissions. The fairness parameter restricts highly reputable clients with a larger number of successful transmissions so that those with fewer successful transmissions can also be selected.

\subsubsection{Client Customization} \label{client_custom}
Another approach to FAFL client selection focuses on customized model settings or customized training procedures. System heterogeneity and statistical heterogeneity are two key challenges in the deployment of FL applications. However, despite the client heterogeneity, current FL paradigms often distribute the same initial models to all clients at the beginning of training rounds. 
As a result, clients with lower capabilities
(in terms of computing hardware, network connections, etc.)
can often be excluded from subsequent rounds. The reason for this is that these clients need more time to complete model training and would be perceived as stragglers by the FL server.
To mitigate this problem, works have been proposed to dynamically adapt the FL model framework or the training procedure based on client capabilities. It promotes fairness by enabling under-represented or never-represented clients to participate in FL.

\begin{figure}[ht]
    \centering
    \includegraphics[width=1\linewidth]{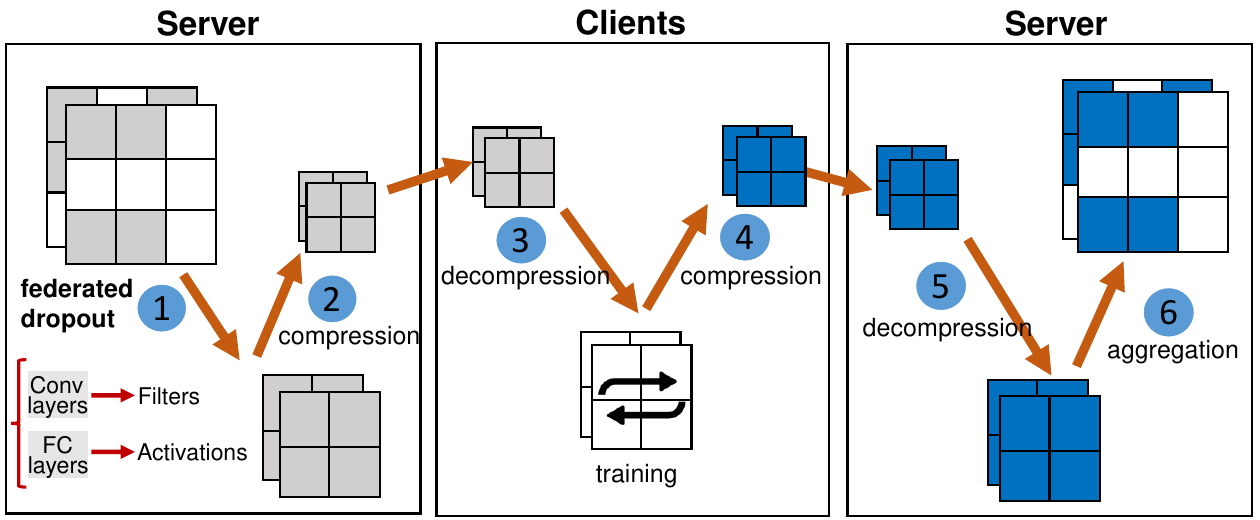}
    \caption{Summary of the Federated Dropout (FD) training procedure \cite{Caldas2018ExpandingTR}. (1) Constructing sub-models using FD, (2) lossy compression before distributing the model to clients, (3) decompression and training on local data, (4) compressing model updates, (5) decompression, and (6) aggregation into the global model.}
    \label{fig:federateddropout}
\end{figure}

\textbf{Customizing Client Models:} \label{F_dropout} Several FAFL works have leveraged the dropout technique \cite{Srivastava2014DropoutAS} to adapt FL models to reduce the bias caused by threshold-based client selection schemes.
In \cite{Caldas2018ExpandingTR}, \textit{Federated Dropout (FD)} was proposed. 
It allows each FL client to receive a sub-model with a size suitable for local training based on its computational resources. The model updates from the clients are then reconstructed by the FL server and aggregated to form the global model.
The training procedures are summarised in Figure \ref{fig:federateddropout}. 
For fully connected layers, a number of activations are often dropped. For convolutional layers, a fixed ratio of filters are often dropped. 
This reduces the computational and communication costs, as smaller sub-models enable more efficient local training and model exchanges between the FL server and the FL clients. 
Lossy compression is applied to further reduce the communication cost.
It is a simple approach that can be applied to CNN-based FL frameworks. By using FD, clients with low capabilities can join FL by training a pruned sub-model. 

While FD successfully decreases communication and local computation costs, it drops activation nodes randomly and does not address the decomposition of the neural networks. It treats them as black-box functions without inspecting the changes in the internal structure of the model as a result of the dropout operations. To address these issues, \textit{Adaptive Federated Dropout} (AFD) was proposed in \cite{Bouacida2020AdaptiveFD}. It maintains an activation score map which is used to select a group of important activation nodes to generate the sub-model that best fits each client, so as to speed up convergence and reduce accuracy loss. 

FD and AFD promote equitable treatment of heterogeneous clients by allowing those with low capabilities to participate in FL training. However, these two approaches have important limitations. Firstly, both of them forward the same pruned model to all clients in a single round, and do not investigate how to provide custom pruned submodels to different clients (i.e., clients with strong capabilities are restricted by those with low capabilities and are not allowed to update more weights). Secondly, FD does not provide the computational benefits at the time of inference because the original model is deployed after training \cite{modelpruneJiang2022}.

To address these limitations, \cite{horvath2021fjord} extended FD to introduce \textit{Ordered Dropout (OD)}, which reduces the computational and memory overhead during training and inference deployment. OD distributes custom-pruned sub-models to clients based on the clients' capabilities. Compared to FD, OD drops
adjacent components of the model instead of random neurons.
Clients with similar computational capabilities are divided into clusters. Those in the same cluster use the same dropout rate. This allows clients with better capabilities to update more weights. For those with weaker memory-related capabilities, the knowledge distillation method \cite{Hinton2015DistillingTK} is used to improve model accuracy by enhancing feature extraction for smaller submodels. OD can also reduce the inference cost, as a client's load can be dynamically adjusted by dropping the least important units at inference time.

\textbf{Customizing Client Workloads:} The idea of customizing the amount of work for an FL client in order to enhance fair client treatment has been explored by current literature. Clients with lower capabilities are assigned less work to enable them to pass threshold-based FL client selection. In \cite{li2020federated}, \textit{FedProx} was proposed to allow partial training to be performed by each client based on its available resources. Instead of assigning a uniform number of epochs to each client, FedProx allows local epochs to vary. This encourages more clients to participate in training, hence reducing the negative impact of system heterogeneity. However, a large number of local updates may lead each client towards the optima of its local objective as opposed to the global objective, resulting in model divergence in the presence of heterogeneous data. Therefore, a proximal term was introduced to restrict the local updates and stabilize the method. 
The proximal term benefits the training process by: (1) encouraging higher-quality local updates by constraining the local updates to be closer to the global model, and (2) safely aggregating local model updates produced generated from different amounts of work. 


\textbf{Customizing based on Client Network Connectivity:}
Client selection can also be affected by a client's communication capabilities (e.g., bit rates, packet loss). When a client sends its local updates to the server during FL, some packets may be lost. Upon detecting a lost packet, the server sends a retransmission request to the client in an attempt to recover the lost packet. However, for clients with slower networks, the retransmission may cause additional delays in FL model training. Thus, model updates from clients with poorer communication channels are less likely to be aggregated into the final model, thereby resulting in model bias.

To handle this issue, authors of \cite{Zhou2021LossTF} proposed a loss-tolerant FL framework, \textit{ThrowRightAway} (TRA). The main idea is that a packet loss may not always be harmful. Based on this assumption, TRA accelerates FL training by ignoring some of the lost packets. In the beginning, all participating FL clients indicate their network conditions to the FL server. Based on their reports, the server classifies the clients into the \textit{sufficient} type and the \textit{insufficient} type. The server then randomly selects the clients to whom to send the global model, and waits for their model updates. When a packet loss is detected, the server only sends a re-transmission request if the client belongs to the \textit{sufficient} type. Otherwise, it simply records the lost packet as zero. After clients finish uploading, TRA recalculates the aggregation based on the packet loss records. The effectiveness of this approach depends heavily on accurately categorizing the clients into different types. It implicitly assumes that the FL clients can accurately assess their own network conditions and will honestly report this information to the FL server. This assumption may not be realistic in practice.

\subsection{Fairness in Model Optimisation} \label{model_optim}

Besides client selection which occurs before FL model training, the optimization process during FL model training can also cause biases in the global FL model. This could cause the model to discriminate against certain protected groups, or to overfit certain clients at the expense of others. As a result, a global model can exhibit inequitable performance across the clients, which can be viewed as unfair treatment by the FL clients. Consider an image recognition scenario in which the FL server has accessed many mobile devices used by the younger users, but only a few used by other age groups. The model may perform very well on the devices of the younger age group, but not so well on devices of other age groups. Recent works have started to investigate the fairness issues in the FL model optimization process. They can be broadly divided into two approaches: 1) objective function-based approaches, and 2) gradient-based approaches. 

\subsubsection{Objective Function-based Approaches}


One common approach to tackle the fairness problem during FL model training is to optimize the global/local objectives of the FL model (e.g., minimising the loss function) so as to satisfy the target fairness constraints. Some existing works with this approach aim to achieve uniform performance across FL clients in terms of prediction accuracy, i.e., performance distribution fairness~\cite{Zafar_2017}. They mainly focus on reducing the variance of model accuracy across clients, while maintaining similar average accuracy. 

\textit{AFL} \cite{Mohri2019AgnosticFL} is one of the earliest works to address fairness in FL training. It aims to achieve the good-intent notion of fairness by preventing the overfitting of the model to any particular client at the expense of others. The authors argue that the distribution over clients, adopted in standard HFL, may not coincide with the target distribution. Hence, under AFL, the global model is optimized for any target distribution formed by a mixture of clients. This does not negatively affect the model performance of other clients as long as they do not increase the loss of the worst-performing client. However, AFL is found to perform well only for a small number of clients. When the number of clients becomes large, the generalization guarantees of the model may not be achievable.

To mitigate the scalability limitation of AFL, \textit{q-FFL} was proposed \cite{Li2020FairRA}, inspired by fair resource allocation methods in wireless communication networks. q-FFL achieves a more uniform accuracy distribution across FL devices, i.e., performance distribution fairness. It achieves this by using a new parameter $q$ to reweigh the aggregate loss by assigning higher weights to devices with higher losses and vice versa. Compared to AFL, q-FFL is more flexible as the degree of fairness can be adjusted by tuning $q$. By setting $q$ to a large value, q-FFL performs similarly to AFL. A more communication-efficient FL aggregation approach based on q-FFL, \textit{q-FedAvg}, was also proposed. q-FFL achieves a lower variance of accuracy (i.e., higher fairness) and faster convergence than AFL. 

Although these approaches promote fairness to a certain extent, they fall short in terms of robustness. For example, both q-FFL and AFL are not robust against adversarial attacks. If a client maliciously increases its loss, it can result in a deterioration in model performance by misleading the global objectives of AFL and q-FFL. To improve the model's robustness while maintaining good-intent fairness, \textit{FedMGDA+} approach was proposed in \cite{Hu2020FedMGDAFL}. It performs a multi-objective optimization by optimizing the loss function of each FL client separately and simultaneously. It uses Pareto-stationary solutions to find a common descent direction for all selected clients so as to avoid sacrificing the performance of any client. Moreover, two techniques: 1) gradient normalization and 2) a built-in robustness method inspired by the Chebyshev approach, are also incorporated to enhance robustness against inflated loss attacks. 
Nevertheless, the accurate identification of malicious clients remains an open problem.

The authors of \cite{Ditto2021} have also addressed the same issue. They proposed a personalized multi-task FL algorithm, called Ditto, which simultaneously improves fairness and robustness. As mentioned earlier, with adversaries, learning globally introduces corruption, while local learning does not generalize well due to the small sample size. Ditto offers a trade-off between these two extremes. After optimizing a global objective function, Ditto allows clients to run finetuning to minimize their individual losses. A regularization term is added to make the personalized models close to the optimal global model. The regularization term in Ditto is similar to the proximal term in FedProx \cite{li2020federated}. The difference is that Ditto learns personalized models, whereas FedProx only learns a global model.


While the aforementioned approaches have generally focused on performance distribution fairness, research works on group fairness are starting to emerge. The authors of \cite{Du2020FairnessawareAF} proposed a fairness-aware method, AgnosticFair, which examines the group fairness of the model by using kernel re-weighting functions to assign a re-weighting value to each training sample in both the loss function and the fairness constraint. It incorporates an agnostic fairness constraint to achieve the demographic parity notion of fairness. AgnosticFair is designed to deal with data shift scenarios \cite{Gama-et-al:2014}. It has been shown to achieve good accuracy and fairness on unknown testing data distribution. Another advantage of AgnosticFair is that fairness can be guaranteed for each local client due to the agnostic fairness constraint, even if the distribution is shifted between the clients and the server. However, prior knowledge is required to determine the re-weighting function. This limits its application in dynamically changing system environments.

The authors of \cite{Cui2021} proposed a multi-objective optimization
framework, FCFL, which achieves good-intent fairness and group fairness simultaneously. Following the idea of AFL, FCFL minimizes the loss of the worst-performing client, leading to a min-max style of fairness guarantees. However, unlike AFL, FCFL utilizes a smooth surrogate maximum function that considers all client objectives instead of using a single non-smooth objective of the worst-performing client. It also adds a fairness constraint on each client to measure disparities across all clients. FCFL employs a gradient-descent-based procedure to find the Pareto solution. The results show that FCFL achieves consistent performance across different clients compared to other methods and also performs significantly better in measuring group fairness. A limitation of FCFL is that it only focuses on local disparity (i.e., client-level group fairness), and it cannot guarantee reasonable global disparity (i.e., group fairness at the global level).

\subsubsection{Gradient-based Approaches}
Compared to the objective function-based approaches, gradient-based approaches are still in the early stages of development. 
The gradient mentioned here refers to clients' local gradient updates in each round. The gradient of client $i$ at iteration $t$ is calculated by $\Delta_i^{(t)} = M_i^{(t)} - M^{(t)}$ where $M^{(t)}$ is a global model and $M_i^{(t)}$ is the updated sub-model after training $M^{(t)}$ on client $i$' data. Some works with this approach aim to compute a fairer average of client updates to achieve uniform performance across clients.

\textit{FedFV} \cite{Wang2021FederatedLW} is one of the earliest works studying fairness guarantees at the gradient level.
It mitigates gradient conflicts \cite{gradientsurgery2020} among FL clients before averaging the gradients in the global objective function. 
The authors argue that the global model could sacrifice model accuracy for some clients for the sake of improving performance for those clients whose gradients differ greatly (These situations often lead to unfairness during FL training).
FedFV uses gradient projection to reduce the conflicts that consist of internal conflicts among selected clients and external conflicts between the selected clients and the unselected clients. 
FedFV has been shown to strike a balance between fairness, accuracy, and efficiency. 
However, the gradient estimation methods applied to mitigate external conflicts might not be consistently reliable because the estimated gradients based on the previous rounds may become out-of-date and incompatible with the latest updates.
Directly applying the estimated gradients to the latest global updates may cause model divergence. A more reliable gradient estimation method that works in conjunction with the FL client selection approach is needed to address this issue.

\subsection{Fairness in Contribution Evaluation}
Contribution evaluation is a sub-field of FL that allows an FL system to assess the importance for the contribution for each client without having access to its local data. The common approach for obtaining this information is through evaluating each client's impact on the aggregated FL model performance. The clients' contribution values may, in turn, be used for client selection and incentive reward distribution in FL. Hence, a fair evaluation of FL clients' contributions is critical. There have been many data valuation methods designed for non-privacy-preserving machine learning settings \cite{Yoon-et-al:2020}. However, they cannot be directly applied under FL settings. Existing FL contribution evaluation approaches can be divided into five categories: 1) self-reported information, 2) individual evaluation, 3) utility game, 4) Shapley value, and 5) empirical methods.

\subsubsection{Self-reported Information}
Here, client contributions are evaluated based on their self-reported information. Such information can be regarding the quality, quantity and collection costs of their local datasets, as well as the computational and communication capabilities they commit to FL.

In \cite{Zhang2020HierarchicallyFF}, data owners are required to report publicly verifiable factors about their local datasets (e.g., data quality, data volume, data collection cost, etc.) to the FL server. The server then uses the reported information to assign ratings to the clients. In \cite{kang2019incentivecontract} and \cite{Ye2020FederatedLI}, the authors applied a similar approach to use self-reported information to build an FL incentive scheme based on Contract Theory \cite{Bolton-Dewatripont:2005}. The server designs contract items and broadcasts them to the data owners. Each item contains rewards and information regarding the clients' local data. Each client selects the most desired items reflecting their contribution types as a commitment to participate in FL. In \cite{motivatingstackberg}, the authors proposed an FL incentive scheme based on a Stackelberg Game \cite{Simaan-Cruz:1973}. The clients' optimal strategy under this game-theoretic setting is to truthfully report their desired prices for one unit of CPU power to the FL task publisher. Self-reported information is also used in auction-based FL incentive mechanism design as such mechanisms allow the data owners to report their costs frequently. In \cite{Le2020AnIM}, the data owners submit their bids which consist of information about the combination of resources, local accuracy, and costs. The FL server uses the bid information to measure each client's potential contribution and then determines the winners.

It should be noted that these methods assume that clients are capable and trustworthy, such that they can reliably assess their own situations and report the information truthfully. In practice, this assumption may not be valid.

\subsubsection{Individual Evaluation}
Individual evaluation approaches measure the contribution of a client based on its performance on specific tasks. It focuses on individual performance instead of the overall performance of the FL model.

A reputation mechanism is widely used to keep track of the historical contribution of an FL participant. The reputation mechanisms are often designed to reflect the participant's reliability and contribution, which can be leveraged by client selection and reward distribution schemes. They can be used in centralized or decentralized FL systems. In \cite{LyuCollaborative2020,LyuRobustFair2020,SongReputation22}, reputation mechanisms were introduced under centralized FL settings. The reputation values are maintained by the FL server. In \cite{LyuCollaborative2020}, the reputation values are updated based on the validation accuracy achieved by each client. In \cite{LyuRobustFair2020}, the authors use cosine similarity between a local model update and the server model update to evaluate the client's reputation.
In \cite{SongReputation22}, the Beta Reputation System \cite{FANG201688,josangbeta2002} is utilized to evaluate the credibility of each client. Based on the value of the loss function, the reputation model classifies client behaviours into positive behaviours or negative behaviours, and then updates the reputation of each client after every round.
For these schemes to work, an accurate and balanced validation dataset is required by the FL server.

In \cite{kangIncentive2019,reliableKang2020,zhang2021IncenRepRvsauction}, reputation mechanisms for decentralised FL systems were proposed.
Each task publisher calculates a client's reputation from two sources: 1) direct reputation opinions from interaction histories with the task publisher, and 2) indirect reputation opinions from other task publishers
The indirect reputation opinions are openly shared and are stored in a blockchain so that no party can tamper with the reputation scores.
In \cite{kangIncentive2019}, the reputation is evaluated by attack detection schemes. Besides, \cite{reliableKang2020} also uses local computation time for reputation evaluation. Both \cite{kangIncentive2019} and \cite{reliableKang2020} utilize a subjective logic model \cite{SubLogic2009} to evaluate reputation.
In \cite{zhang2021IncenRepRvsauction}, reputation is measured based on the clients' local model gradient updates in each round. 
Each reputation evaluation method has its own limitations. 
Both \cite{kangIncentive2019} and \cite{reliableKang2020} rely on Assumption 2a (i.e., the clients are trustworthy) which may not always be realistic. For example, in \cite{reliableKang2020}, the task publisher uses the local computation time and dataset size to estimate the proportion of local data the client has dedicated to training the FL model. This method is susceptible to cheating by misbehaving clients who intentionally lengthen their local computation times so as to appear to be using a large proportion of local data for training. For \cite{zhang2021IncenRepRvsauction}, the task publisher needs to save all local models and the global model during all historical rounds of training. This can result in a high storage overhead as the number of clients and/or the number of training iterations increase.

Besides reputation, there are other approaches for evaluating a FL client's individual contribution. Zeng et al. \cite{Zeng2020FMoreAI} proposed a score function based on each client's bid and resource quality under an auction-based FL client selection scheme. The top-scoring clients are selected to participate in FL. In \cite{Lyu2019TowardsFA}, the authors proposed a mutual evaluation-based approach for pairs of FL participants to assess each other's potential value. Such assessments leverage data generated based on each participant's local data, and protected through local differential privacy \cite{Cormode-et-al:2018}.

Two assumptions are often adopted in individual evaluation approaches: 1) both the FL server and the FL clients are trustworthy; and 2) a participant with a local model similar to models from other participants (or with the global model) is deemed to provide more contribution. These two assumptions might not always hold in practice. For Assumption 1, it is well-known that the FL server and the FL clients may be selfish and misbehave. For Assumption 2, under non-i.i.d.\footnote{“non-i.i.d” means data are not identically distributed. More precisely, the data
distributions of different clients may be different from each other} settings, participants usually hold datasets with heterogeneous class distributions. In such cases, dissimilar model updates from participants with different data distributions can provide valuable complementary knowledge for improving FL model performance. These factors, if not handled properly, negatively affect the perceived fairness of contribution evaluation.

\subsubsection{Utility Game}
Utility game-based FL contribution evaluation approaches are closely related to profit-sharing schemes \cite{gollapudi_et_al2017} -- rules that map the utility produced by the participants into their corresponding rewards. There are three widely adopted profit-sharing schemes: 
\begin{enumerate}[leftmargin=*]
    \item Egalitarian: any unit of utility produced by a team is divided equally among the team members; 
    \item Marginal Gain: the payoff of a participant is equal to the utility that the team gained when he/she joins; and
    \item Marginal Loss: the payoff of a participant is equal to the utility lost when he/she leaves the team. 
\end{enumerate}
For marginal gain and marginal loss, the payoff amount received by each client depends on its order of joining, as payoff schemes generally aim to motivate clients to join as early as possible.

The most commonly used scheme in FL is the marginal loss scheme. Wang et al. 2019 \cite{wangguanContribution} adopted the marginal loss approach to measure the contributions of different parties in HFL. They use an approximation algorithm to implement the influence measures. Nishio et al. 2020 \cite{Nishio2020EstimationOI} adopted the marginal loss scheme to evaluate the contribution for each client during a single FL training process in order to reduce communication and computation overhead. These approaches aim to achieve the notion of contribution fairness. 

Simple marginal loss schemes are suitable for fairly evaluating the contribution of a given client among a given set of clients who are collaboratively training an FL model. This is a relative evaluation (i.e., it depends on how much the other participating clients contribute), and does not reflect the actual value of the client's local data.
Shapley value has been leveraged to address this shortcoming.

\subsubsection{Shapley Value}
Shapley Value (SV)-based FL contribution evaluation approaches have attracted much research attention in recent years. SV, a marginal contribution-based scheme, was introduced in 1953 as a solution concept in cooperative game theory \cite{shapley1997value}. Consider $n$ clients with data sets $D_1, D_2, \dots, D_n$, a machine learning algorithm \(\mathcal{A}\), and a standard test set $T$. $D_S$ is a multi-set, where $S \subseteq N = \{1, 2, \dots, n\}$. A model trained on $D_S$  through algorithm \(\mathcal{A}\) is denoted by $M_S(\mathcal{A})$ which is abbreviated as $M_S$. The performance of model $M$ evaluated on the standard test set $T$ is denoted by $U(M,T)$, abbreviated as $U(M)$. The Shapley value $\phi(\mathcal{A}, D_N, T, D_i)$, abbreviated as $\phi_i$, can be used to calculate the contribution of each FL client $i$:
\begin{equation} \label{SVequation}
\phi_i = C  {\sum_{S \subseteq N \setminus \{ i\}} \frac{U(M_{S\cup \{i\}}) - U(M_S)}{\bigl(\begin{smallmatrix} n-1 \\ |S|  \end{smallmatrix}\bigr)} }.
\end{equation}
where $C$ is a constant.

By averaging the sum of the marginal contribution over all subsets of $D$ not containing $i$,
SV reflects $i$'s contribution to the FL model as a result of only its local data, regardless of its order of joining the coalition. In this way, it can produce fairer client contribution evaluation.
Nevertheless, the computational complexity of calculating SV is $O(2^n)$, which is exponential. To improve the efficiency of SV calculation, many heuristic methods have been proposed in traditional machine learning (e.g., Truncated Monte Carlo Shapley (TMC-Shapley) and Gradient Shapley \cite{pmlr-v97-ghorbani19c}). Inspired by these approaches, SV-based FL client contribution evaluation approaches are emerging.

\textbf{Client-based Approaches:}
Song et al. 2019 \cite{Song9006327} proposed two gradient-based SV methods: 1) One-Round Reconstruction (OR) and 2) Multi-Round Reconstruction (MR). Both methods gather gradient updates from FL clients to reconstruct the FL model, instead of retraining with different subsets of clients. OR gathers all the gradient updates over all the training rounds. Then, it reconstructs the models for all the subsets in the final round. OR calculates SV only once by using the reconstructed models in the final round. In contrast, MR calculates a set of SVs in each round of training, and then aggregates them to compute the final SV-based contribution values. Wei et al. 2020 \cite{WeiDataValuation2020} extended MR to propose the Truncated Multi-Round (TMR) method. TMR improves on MR in two ways. Firstly, it assigns higher weights to the training rounds with higher accuracy values. Secondly, TMR improves efficiency by skipping the last few rounds.

By leveraging these gradient-based SV estimation methods, the efficiency of evaluating FL client contribution can be significantly improved. However, we still need to evaluate the sub-models for different subsets of clients in each round of training. To further reduce the computational cost, Wang et al. 2020 \cite{WangDataValue2020} proposed two efficient approximation approaches to improve the efficiency of within-round SV calculation, inspired by \cite{pmlrv89jia19a}: 1) permutation sampling-based approximation, and 2) group testing-based approximation. 
The authors of \cite{GTGshapley} proposed the Guided Truncation Gradient Shapley (GTG-Shapley) approach which combines between-round and within-round truncation to further reduce the training cost. Between-round truncation eliminates entire rounds of SV when the remaining marginal gain is small. Within-round truncation skips the remaining sub-model evaluation in permutations when the remaining marginal gain is small.

\textbf{Feature-based Approaches:} Vertical Federated Learning (VFL) \cite{FL:2019} in which participants' datasets share little overlap in the feature space but significant overlap in the sample space, raises new challenges for contribution evaluation. In \cite{wangguanContribution}, Shapley value is leveraged to calculate the feature importance in VFL. Since directly using SV to evaluate each prediction could reveal the potentially sensitive features, the authors proposed to perform SV calculation on groups of features instead of on each individual feature. However, this method is still computationally expensive since the computational cost increases exponentially with the training data size.

\subsubsection{Empirical Methods}
Data-based counterfactual contribution evaluation, such as Shapley Value, has produced promising results for FL contribution evaluation. However, their high computational costs and imprecision as a result of estimation to enhance efficiency still limit their scalability. Empirical methods for contribution evaluation have been studied as alternatives to theory-based FL client contribution evaluation. Shyn et al. 2021 \cite{Shyn2021FedCCEAA} proposed FedCCEA, which learns the data quality of each client by constructing an Accuracy Approximation Model (AAM) with sampled data size. This method robustly and efficiently approximates the client's contribution by using the sampled data size, and allows partial participation by clients through setting the desired sizes of local data to be used for FL model training. FedCCEA consists of a simulator and an evaluator.
The simulator obtains the inputs (i.e., sampled data size) and the targets (i.e., round-wise accuracy) of AAM by running one-epoch FL classification tasks. Then, the evaluator optimizes the weight vectors from AMM by using the stored inputs and targets. After model convergence, the shared weights of the first layer are extracted to learn the importance for data size for each client. 
Nevertheless, since AAM is built upon a very simple neural network architecture, FedCCEA is currently limited to simple FL tasks, making it less well-suited for practical applications. 

\subsection{Fairness in Incentive Mechanisms}
Existing works on FL usually assume that FL clients are always willing to join FL when invited. However, in reality, FL clients may be reluctant to join the training, as 1) clients need to contribute their computational/communication resources, and 2) the malicious attackers may still infer the private information of training data from the local gradients. Such training costs and security risks hamper clients' continued participation.
Imagine when an FL platform invites companies from the same business sector for the FL tasks. The companies are generally reluctant to join FL since contributing their data to train a federated model that is subsequently shared with potential competitors can incur significant opportunity costs to a company. 
Hence, fair incentive mechanisms, which provide satisfactory compensation arrangements, should be designed for FL to encourage companies (a.k.a, data owners) to participate in collaborative model training.
A fair incentive mechanism distributes "fair" rewards to FL clients based on their contributions and training costs. For higher accuracy of the client model updates, there will be an increase in the reward for the corresponding clients, such that the system can attract high-quality clients to join the training continually.
Many incentive mechanisms for FL have been proposed in recent years \cite{zhan9369019}. Not all of them focus on enhancing fairness. For example, in Stackelberg game-based FL scenarios \cite{Feng-et-al:2018,Sarikaya-Ercetin:2019}, the server and the clients compete to optimize their own utilities. The equilibrium solution of the game achieves a trade-off between the two parties, but might not be fair. In this section, we focus on FL incentive mechanism designs that take fairness into consideration.
Different approaches and fairness criteria have been proposed. We classify these incentive mechanisms into two categories: 1) Monetary Incentive Mechanisms, which distribute monetary payoffs to FL clients; and 2) Non-Monetary Incentive Mechanisms, which use FL models with different levels of performance to incentivize FL clients.

\subsubsection{Monetary Incentive Mechanisms}

Zeng et al. 2020 \cite{Zeng2020FMoreAI} proposed FMore, which extends the multi-dimensional procurement auction from \cite{auction1993che} to motivate more high-quality data owners to join FL while minimizing total cost. Although it uses game theory to derive the optimal strategy for distributing profits among the clients (which accounts for competition among data owners), it takes fairness into consideration during contribution evaluation. In the bidding stage, the server broadcasts its asking price with a scoring rule to the clients
\begin{equation}
    S(q_{i1},q_{i2},\dots,q_{im},p_i) = s(q_{i1},q_{i2},\dots,q_{im}) - p_i
\end{equation}
where $s(\cdot)$ is the utility function of the FL server. $q = (q_1,\dots,q_m)$ is the quality vector of clients' resources (i.e., local data, computation capability, bandwidth, CPU cycle, etc.). $p_i$ is the expected payment for $i$. After receiving an ask, the clients decide on whether to bid or not based on their available resources.
After collecting sufficient bids, the server selects clients based on their estimated contribution values. By applying the same scoring rule, the clients can determine whether they are being treated fairly. This may further encourage clients' participation. 

One shortcoming of FMore is that it does not consider fairness in client selection. Hence, to tackle this issue, the authors further proposed an extension of FMore, $\psi$-FMore, which accounts for fairness in client selection. $\psi$-FMore assigns a probability $\psi$ to each client, which is used to increase the winning probability of low-score clients and decrease the winning probability of high-score clients as situations change. However, this work assumes that both the server and the clients bid truthfully. In the presence of misbehaving FL participants, the training performance deteriorates. Moreover, privacy preservation needs to be enhanced in the score calculation stage to avoid revealing sensitive information. 

Besides accurately evaluating the contribution of each client, a fair incentive mechanism also needs to ensure that each client is paid fairly based on its contribution to the FL model (referred as contribution fairness). Cong et al. 2020 \cite{cong2020vcgbased} proposed Fair-VCG (FVCG) based on the VCG mechanism. FVCG incentivizes the data owners to truthfully report their costs and data qualities. 
Then, the server distributes the rewards to all the data owners by setting the same unit price for data quality for all data owners. However, as aforementioned, the truthfulness of self-reported information cannot be guaranteed in practice. 

To address this limitation, Zhang et al. 2021 \cite{zhang2021IncenRepRvsauction} leveraged reputation to reflect the quality and reliability of the data owners indirectly, instead of asking them to self-report. It proposed the RRAFL mechanism based on reputation and reverse auction for decentralized FL systems. The FL task publisher is responsible for keeping track of clients' historical behaviors to derive direct reputation value. In addition, reputation records from multiple FL task publishers can be shared to obtain indirect reputation values. The task publisher sorts the FL participants based on their unit reputation bid prices, and selects the $k$ participants with the lowest unit reputation bid prices to join FL. Then, the task publisher uses the unit reputation bid price of the $(k + 1)$-th participant to determine the payoff for each selected participant, since the final unit bid price provided by the task publisher would be higher than the unit bid prices of all the selected participants. As a result, the selected participants can be effectively motivated to join FL.
However, reputation-based approaches rely on the availability of historical records of FL client performance. The issue of boot-strapping the reputation system in FL needs to be addressed in order for these approaches to operate.

Contract theory \cite{Bolton-Dewatripont:2005} has also been adopted to design FL incentive mechanisms. In mobile networks, there exists information asymmetry between the task publisher and the mobile devices (i.e., the task publisher does not know the data quality, data quantity and available computational resources of each mobile device). It may incur a high cost if it were to monitor such information. To reduce the impact of information asymmetry, Kang et al. 2019 \cite{kang2019incentivecontract} applied Contract theory to design an efficient incentive mechanism to attract mobile devices with high-quality data to join FL. It defines a parameter of local data quality as the type of the contract model. The task publisher uses the observations in the previous round to design different contracts for data owners with different data qualities. As a result, devices with higher quality data and more computational resource contribution receive higher rewards from the task publisher. Similarly, \cite{Ye2020FederatedLI} also used Contract theory to design an incentive mechanism for FL in vehicular edge computing settings. It introduces a 2-dimensional contract to determine the appropriate rewards based on clients' data quality and computational capabilities. One shortcoming of these approaches is that they assume a monopoly market with only one FL task publisher. As the monopolist, the task publisher only provides limited contract choices to each device, which negatively impacts the mobile devices' profits. In practice, there could be many FL task publishers competing to attract FL clients. Currently, such scenarios have not yet been adequately accounted for by existing FL incentive mechanisms.

The above-mentioned FL incentive mechanisms implicitly assume that the incentive budget has been pre-determined. There are situations in which the incentive budget is not available at the time of FL model training. Instead, the participants expect to be rewarded with the revenue generated by the FL model at a later time \cite{kairouz2021advances}.
To account for this scenario, Yu et al. 2020 \cite{Yu2020AFI} defined two notions of fairness criteria that are important to the long-term sustainable operation of FL in addition to contribution fairness: regret distribution fairness and expectation fairness. Regret distribution fairness requires that clients be treated fairly based on their waiting time for receiving their incentive rewards. Since the training and commercialization of the FL models take time, the server may not have enough revenue to compensate the participants at the early stage. This leads to a temporary mismatch between clients' contribution and the rewards they hitherto receive. To overcome this issue, \cite{Yu2020AFI} proposed a dynamic payoff-sharing scheme - Federated Learning Incentivizer (FLI) - which dynamically distributes the rewards to the clients by maximizing the collective utility and minimizing inequality between data owners' regret and waiting time. It ensures that the clients, who have contributed more high-quality data and have waited longer for the full payoff, will receive more revenues in the subsequent rounds. During the gradual payout of clients' rewards, the notion of expectation fairness is used to ensure that the clients' regret values are reduced as equitably as possible in lieu of their contributions.


\subsubsection{Non-Monetary Incentive Mechanisms}
Another category of FL incentive mechanisms is not based on monetary rewards. Instead, they seek to motivate clients by assigning them FL models with different performances based on their contributions. These approaches are suitable for application scenarios in which 1) a monetary incentive budget is not available, 2) leveraging future revenues generated by the FL model to reward the clients is not feasible, or 3) there is competition among FL clients which causes high contribution clients to feel unfairly treated if they receive the same FL model as low contribution clients (i.e., the free-rider problem \cite{kairouz2021advances}).

Zhang et al. \cite{Zhang2020HierarchicallyFF} proposed the Hierarchically Fair Federated Learning (HFFL) framework. HFFL ensures fairness among the clients by providing the clients with higher-quality data with higher quality model updates. HFFL first classifies the clients into different levels based on their characteristics (e.g., data quality, data size). Then, it trains multiple FL models - one for each level. To train a lower-level model, clients from higher levels only contribute the same amount of data as those from lower levels. On the other hand,  clients from lower levels are required to contribute all their local data when collaboratively training a higher-level FL model. As a result, clients at higher levels receive FL models with better performance. However, HFFL has several shortcomings. Firstly, the clients at the same level may not have the same amount of data since additional factors are being considered when categorizing clients. Secondly, some of the client characteristics used for categorizing clients are obtained through self-reporting, which makes the approach susceptible to fake reporting.

In contrast to HFFL, which divides all the clients into different clusters and trains one model for each cluster, \cite{Lyu2019TowardsFA} proposed a decentralized Fair and Privacy-Preserving Deep Learning (FPPDL) framework in which each participant receives a different variant of the final FL model based on his contribution. In FPPDL, each participant earns a certain number of transaction points based on their local credibility and commitment level. The local credibility system is maintained by mutual evaluation between any two participants. Then, participants can use their transaction points to download gradients from other participants. As a result, each participant can obtain an improved local model compared to their standalone models without collaboration, and the improvement gained by each participant is proportional to his corresponding contribution. Moreover, FPPDL also preserves data privacy by incorporating Differentially Private GAN (DPGAN) and a three-lay onion-style encryption scheme. 
Nevertheless, the current approaches still lack precision control when it comes to constructing a machine learning model with a specific level of performance.

\section{FAFL Evaluation Metrics}
For the long-term sustainability of FAFL, it is important to establish a set of performance evaluation metrics so that the relative advantages of various proposed approaches can be objectively assessed. 
In this section, we discuss some common evaluation metrics adopted in current FAFL research. 

\begin{itemize}[leftmargin=*]
    \item \textbf{Accuracy}: Accuracy is used as the performance metric for client selection \cite{li2020federated, Huang2021AnEC, HuangCS2022}, training optimization \cite{Li2020FairRA, Mohri2019AgnosticFL, Hu2020FedMGDAFL, Wang2021FederatedLW}, and reward distribution \cite{Zhang2020HierarchicallyFF}. Training loss \cite{li2020federated, Hu2020FedMGDAFL, Zeng2020FMoreAI} has also been used to measure FL model performance related to accuracy. In FAFL, most works adopt average validation accuracy and accuracy variance to evaluate the performance of the FL model \cite{Wang2020OptimizingFL, Hu2020FedMGDAFL, Wang2021FederatedLW, Li2020FairRA}. Nevertheless, some works have also used test accuracy to evaluate the performance of the FL model on each FL client \cite{LyuCollaborative2020, Khan2020FederatedLF, pandey2020crowdsourcing, Hu2020FedMGDAFL, Wang2021FederatedLW, Shyn2021FedCCEAA}. Test accuracy has the following advantages: 
    1) It avoids sacrificing the performance at any client for the sake of improving the overall performance; and
    2) Clients can be rewarded accordingly to their contributions.
    
    \item \textbf{Efficiency}: Efficiency is another commonly used FL performance metric. Many works measure training efficiency by tracking the total time used for model training \cite{Song9006327, Nishio2019ClientSF, GTGshapley} or the number of FL training rounds \cite{Huang2021AnEC, Li2020FairRA, HuangCS2022, Bouacida2020AdaptiveFD, Caldas2018ExpandingTR, Lyu2019TowardsFA, Wang2021FederatedLW}. Some works have also tracked the training time across FL training rounds to evaluate the reduction in training time in order to measure efficiency \cite{Huang2021AnEC, Zeng2020FMoreAI}. Note that efficiency is not used to evaluate model performance, but to evaluate the performance of the training framework.
\end{itemize} 

In addition to performance evaluation metrics, the following fairness-related metrics have been adopted by FAFL research. 
\begin{enumerate} [leftmargin=*]
    \item \textbf{Average Variance}: The Average Variance (AV) is usually applied to measure the performance variation of an algorithm across different devices. It is used to quantify the degree of fairness during optimization \cite{Li2020FairRA, Hu2020FedMGDAFL, Wang2021FederatedLW}.
    
    AV has been extended to FL setting as follows: 
    \begin{equation}
        AV = \frac{1}{n} \sum_{i=1}^n (F_i(t) - \overline{F}(t))^2
    \end{equation}
    where $\overline{F}(t)$ refers to the average accuracy across all FL clients. 
    A lower AV value indicates a higher degree of fairness for a given FL method. Let $\{F_1(t), \dots, F_n(t)\}$ and $\{F_1(t'), \dots, F_n(t')\}$ be the accuracy distribution among $n$ clients for two FL methods $t$ and $t'$, respectively.
    We say that method $t$ is fairer than $t'$ if $AV (F_1(t), \dots, F_n(t)) < AV (F_1(t'), \dots, F_n(t'))$. 
    Some research works \cite{Ditto2021} use the standard deviation of performance across different devices to measure fairness. Since both standard deviation and AV share the same idea of reflecting the variability in distribution (i.e., standard deviation is the square root of AV), we also include these works in this section.
    
    There are some drawbacks of using AV to evaluate algorithmic fairness: i) it is sensitive to outliers; and ii) it only measures the relative fairness by assigning a higher degree of fairness to the method with lower AV. 
    
    \item \textbf{Distance Metrics}: Distance metrics (e.g., Cosine Distance, Euclidean Distance, Maximum Difference, Mean absolute error) have been applied to measure the similarity between the performance of various FL methods in order to study their fairness. They are commonly adopted by FL contribution evaluation methods \cite{Song9006327, WeiDataValuation2020, GTGshapley} to measure how accurate the estimated FL participants' contributions are. Let ${\phi}^* = \langle {\phi_1^*},{\phi_2^*},\dots,{\phi_n^*} \rangle$ and ${\phi} = \langle {\phi_1},{\phi_2},\dots,{\phi_n} \rangle$ be the vectors of normalized contributions for $n$ clients calculated by two methods $t^{*}$ and $t$ based on Eq. \eqref{SVequation}. Normalization is performed to remove numerical differences and reflect the ratio difference between the contributions of two clients. Here, $t^{*}$ is a method that can compute the ground truth contributions by the FL participants.
    \begin{enumerate}
        \item The \textbf{Euclidean Distance} between $\phi^*$ and $\phi$ is: 
        \begin{equation}\label{euclidean}
        D_E = \sqrt{\sum_{i=1}^n ({\phi_i^*} - {\phi_i})^2}.
        \end{equation}
        The disadvantages in using Eq. \eqref{euclidean} are:    
        1) It is not scaled in-variant (i.e., ${\phi_i^*}$ and ${\phi_i}$ need to be normalized before being applied); and 2) It does not perform well in high-dimensional spaces due to the curse of dimensionality\footnote{The points essentially become uniformly distanted from each other in high-dimensional spaces.} \cite{distancemetrics2001}. Hence, it may not be suitable when there are a large number of clients.
        \item The \textbf{Manhattan Distance} is computed as: 
        \begin{equation}
        D_M = \sum_{i=1}^n |{\phi_i^*} - {\phi_i}|.
        \end{equation}     
        $D_M$ measures the sum of the absolute differences between the two vectors. It is more suitable for high-dimensional data than the Euclidean distance  \cite{distancemetrics2001}.
        \item The \textbf{Cosine Distance} is computed as:
        \begin{equation}
        D_C = 1 - \cos({\phi^*,\phi})
        = 1 - \frac{\sum_{i=1}^n {\phi_i^*}\times {\phi_i}}{ \sqrt{\sum_{i=1}^n {\phi_i^*}^2} \times \sqrt{\sum_{i=1}^n {\phi_i}^2}}.
        \end{equation}
        $D_C$ measures the similarity in angles of vectors ${\phi_i^*}$ and ${\phi_i}$, but not in their magnitude \cite{SimArash2016}.
        \item The \textbf{Maximum Difference} is computed as:
        \begin{equation}
        D_{Max} = \max_{i=1}^n |{\phi_i^*} - {\phi_i}|.
        \end{equation}
        $D_{Max}$ represents the greatest difference between two vectors along any coordinate dimension. Unlike Euclidean distance and Cosine distance, Maximum distance is typically used in specific use-cases.
        In \cite{Song9006327}, it is used to measure the maximum percentage difference between the performance of a client's local model and the FL model.
        \item The \textbf{Mean Absolute Error (MAE)} is:
        \begin{equation}
        MAE = \frac{\sum_{i=1}^n|{\phi_i} - {\phi_i^*}|}{n}.
        \end{equation}
        \item The \textbf{Root Mean Squared Error (RMSE)} is:
        \begin{equation}
        RMSE = \sqrt{\frac{\sum_{i=1}^n({\phi_i} - {\phi_i^*})^2}{n}}.
        \end{equation}
    \end{enumerate}
    MAE and RMSE are both commonly used to evaluate errors. 
    They have been adapted to measure the distance between an FL client's contribution calculated by a proposed method and the ground truth contribution value.
    
    \item \textbf{Pearson Correlation Coefficient}: The Pearson Correlation Coefficient (PCC) with the actual Shapley value computed by the canonical Shapley value formulation has been used to measure the fairness of contribution evaluation methods. Let ${\phi_i}$ denote client $i$'s Shapley value calculated by a given approximation method, and ${\phi_i^*}$ denote client $i$'s ground truth Shapley value. $\overline{\phi^*}$ and $\overline{\phi}$ are means of values of ${\phi^*}$ and $\phi$ respectively. Let $s_{\phi_i^*}$ and $s_{\phi_i}$ represent the corresponding standard deviations. PCC is computed as:
    \begin{equation}
    PCC = \frac{\sum_{i} ({\phi_i^*} - \overline{\phi^*})({\phi_i} - \overline{\phi})}{{s_{\phi_i^*}} \times {s_{\phi_i}}}.
    \end{equation}
    The higher the PCC value achieved by a given method, the fairer it is in terms of evaluating an FL participant's contribution to the FL model. The drawbacks of PCC are: i) it is sensitive to outliers; and ii) the result may be inaccurate if $\phi$ and $\phi^*$ are not linearly related since PCC assumes that the two data variables have a linear relationship.
    
    \item \textbf{Jain's Fairness Index (JFI)}: Jain’s index \cite{Jain1998AQM} is a fairness measure widely used in computer networks and resource allocation (e.g., to identify underutilized communication channels). 
    In \cite{divi2021new,Yae2020,Abdelmoniem2021OnTI}, researchers adapted JFI to evaluate the fairness in terms of uniform local performance across clients.
    It is computed as: 
    \begin{equation} \label{jain1}
    JFI = \frac{({\sum_{i=1}^n {F_i(t)})^2}}{n \times \sum_{i=1}^n ({F_i(t)})^2}
    \end{equation}
    where ${F_i(t)}$ is the local objective function of client $i$. The values of $JFI$ range from $\frac{1}{n}$ (i.e., most unfair) to 1 (i.e., most fair) where 1 means that all clients have the same performance (i.e., following performance distribution fairness). JRI is flexible enough to accommodate a variety of notions of fairness.  ${F_i(t)}$ can also be replaced by the amount of information received by client $i$ \cite{Liu2022AFA} or the number of times of participation by client $i$ to quantify selection fairness.
    
    Besides Eq. \eqref{jain1}, JFI can also be formulated as:
    \begin{equation}
    JFI = \frac{({\sum_{i=1}^n \frac{x_i}{z_i}})^2}{n \times \sum_{i=1}^n (\frac{x_i}{z_i})^2}
    \end{equation}
    where $\frac{x_i}{z_i}$ is a normalized value about a given client $i$, which can represent any characteristics related to how the FL scheme treats $i$ (e.g., $\frac{x_i}{z_i}$ can be $i$'s reward or $i$'s times of participation $x_i$ divided by its contribution index $z_i$). In this way, JFI can be further adapted to evaluate contribution fairness.
    JRI is more advantageous compared to the aforementioned metrics, which can only measure the relative fairness of different FL approaches, but not how close they are to absolute fairness. 
    
    \item \textbf{Risk Difference Metrics}: Risk difference measures the disparity between positive predictions on the sensitive group and those on the non-sensitive group, which is used to quantify group fairness. In \cite{Du2020FairnessawareAF, Cui2021}, the risk difference of a classifier $f$ is measured by two metrics:
    \begin{enumerate}
        \item The marginal-based metric \textbf{Demographic Parity} \cite{demographic2012}:
        \begin{equation}
        DP(f) = |P(\hat{Y}=1|A=0)-P(\hat{Y}=1|A=1)|
        \end{equation}
        \item The conditional-based metric \textbf{Equal Opportunity} \cite{EqualOpp2016}:
        \begin{equation}
        OE(f) = |P(\hat{Y}=1|A=0,Y=1)-P(\hat{Y}=1|A=1,Y-1)|   
        \end{equation}
    \end{enumerate}
    where $\hat{Y}$ is the predicted value of $f$, $A$ is the binary sensitive attribute, and $Y$ is the true target outcome. A low risk difference value implies a high degree of fairness.
\end{enumerate}

\begin{figure*}[t!]
    \centering
    \includegraphics[width=1\linewidth]{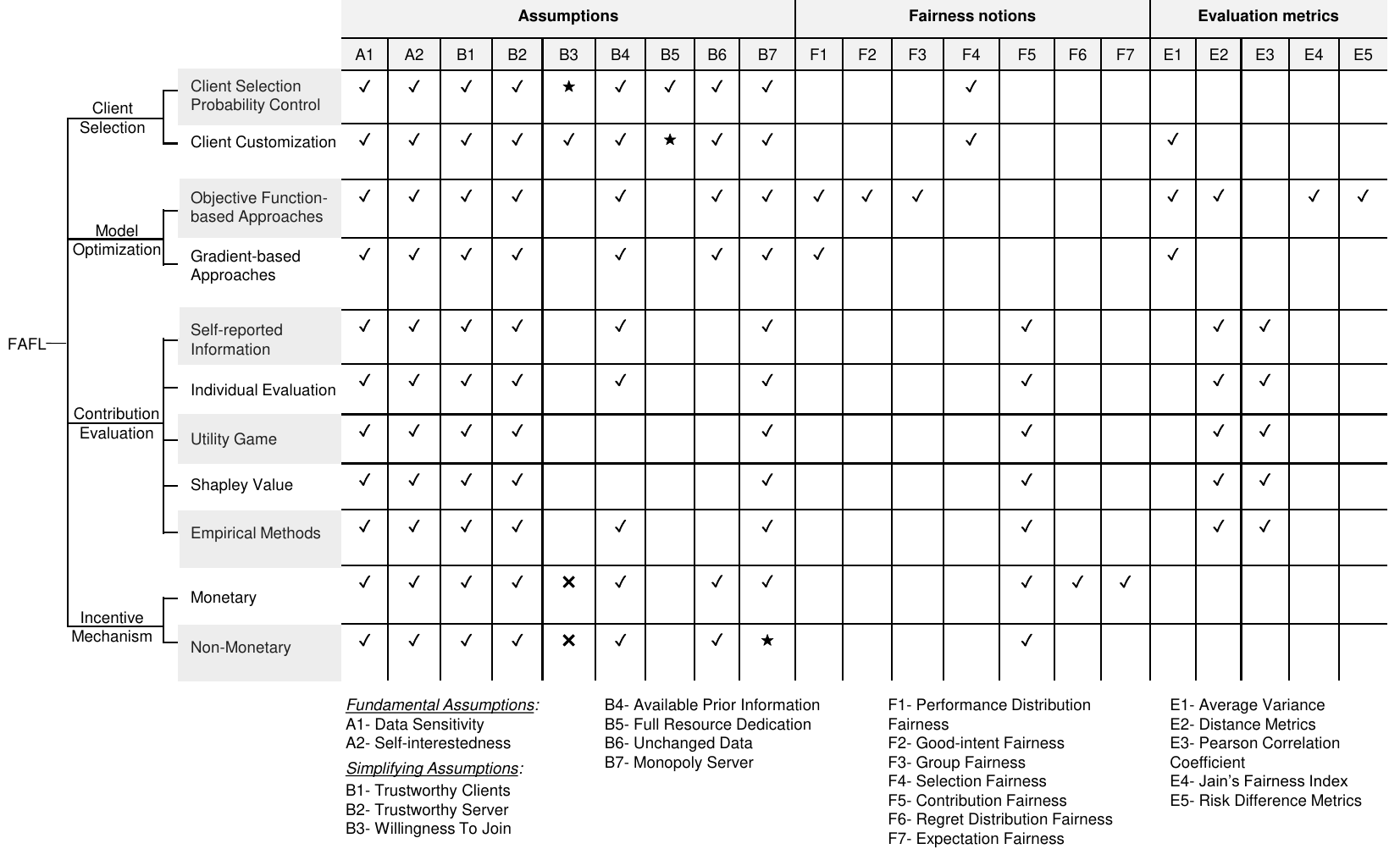}
    \caption{Types of Assumptions, Fairness notions and Evaluation metrics used in different FAFL approaches. (1) The Tick symbol means that it is commonly adopted by the given approach, (2) the Cross symbol means that it is avoided by the given approach, and (3) the Star symbol means that some studies in the given approach are starting to relax the given assumption. The blank cells indicate that it is not applicable to the given approach.}
    \label{fig:ticktable}
\end{figure*}

\section{Promising Future Research Directions}
FAFL research is still in its infancy. Much research is still needed to strike a balance between FL server and the FL client interests. 
In Figure \ref{fig:ticktable}, we summarize the proposed taxonomy together with the assumptions, notions of fairness and fairness evaluation metrics our review has uncovered. It can be observed that the fundamental assumptions (A1--A2) are adopted in most FAFL approaches. For simplifying assumptions, the FL incentive mechanism research commonly rejects assumption B3 (i.e., data owners always agree to join FL when invited). Assumption B5 (i.e., a client dedicates all its resources to a given FL training task) is being relaxed in client customization research under FAFL. In the Non-Monetary Incentive Mechanism domain, there are works starting to emerge that relax assumption B7 (i.e., monopoly market structure). Among the notions of fairness, notion F5 (i.e., contribution fairness) is most widely adopted in contribution evaluation and incentive mechanism research. Notion F1 (i.e., performance distribution fairness) is commonly adopted in model optimization research, while notion F4 (i.e., selection fairness) is widely adopted in client selection research. Nevertheless, these notions of fairness are not jointly addressed in any one approach. There is a diversity of opinions among researchers in this area. One important finding in the usage of evaluation metrics is that the most relevant evaluation metric - Jain's Fairness Index (metric E4) - is capable of evaluating diverse notions of fairness. While most works focused on applying JFI to Notion F1 (i.e., performance distribution fairness), future work can further adopt JFI to evaluate other fairness notions. In addition to the aforementioned potential areas of improvement, we highlight six promising future research directions in FAFL.

\subsection{Temporal consideration in contribution evaluation}
Existing FL contribution evaluation works mainly focus on evaluating data quality, data collection cost or improvement in model performance. However, recent research has pointed out that there is a temporary mismatch between contributions and rewards \cite{Yu2020AFI}. In scenarios where the incentive for FL participants comes from future revenues generated by the resulting models, FL clients' waiting time for receiving the full payoff needs to be accounted for fair treatment. These temporal costs incurred by FL participants need to be taken into account when designing holistic FL contribution evaluation schemes. 

\subsection{Incentive mechanism for non-monopoly FL settings}
In current FL incentive mechanism design research, most works assume a monopoly market setting consisting of only one FL task publisher and multiple FL clients. The task publisher distributes the rewards to the clients to incentivize them to participate in the federated model training. Under this assumption, the single task publisher faces no competition (i.e., clients can only choose to join this federation or not with no other alternatives). 
Such an assumption is not realistic and hinders the development of more FAFL incentive schemes.
A non-monopoly market setting deserves further study as it involves many task publishers. The task publishers compete with each other to attract more clients. In such a scenario, a task publisher faces stronger pressure to provide more reasonable rewards to motivate clients to join federated training. This may inspire fairer FL incentive schemes to emerge.

\subsection{Deterrence mechanism design}
As mentioned in Section \ref{ComA}, existing FAFL approaches mostly assume that FL clients are trustworthy. However, in reality, there exist malicious clients who might tamper with the gradients and other information they pass to the FL server. This can severely degrade model performance. Although there have been works focusing on defending against privacy attacks, they generally overlook fairness. For example, some methods may filter out rare but informative updates from clients with minority classes, leading to unfair treatment of such clients and potentially biased FL models \cite{NEURIPS2020_b8ffa41d}. Hence, more work is needed to study attacks and defense models to design robust approaches towards fairness in FL. One potential direction is to establish a penalty system to deter misbehaving clients. For example, Stackelberg game-based approaches \cite{Simaan-Cruz:1973} can be leveraged to dynamically inspect client local model updates for potential attacks and design corresponding punitive measures. The key is to ensure that the expected utility gain from an attack is negative so as to deter any rational attacker. 

\subsection{Model performance control for non-monetary incentive}
In model-based non-monetary FL incentive mechanisms, the clients are rewarded with different versions of models having performance commensurate with their contributions. If Client A contributes more than Client B, Client A will receive a final model with better performance. However, existing approaches cannot precisely control the performance of the different variants of the FL model allocated to each client. In order for such schemes to be adopted in practice, this research challenge needs to be overcome.

\subsection{Social norm-based federation formation}
Existing FL client selection approaches are mostly designed from the perspective of the FL server. The monopoly assumption implies that clients are not able to choose alternative servers. In order to support the emergence of more realistic FL-based data exchange marketplaces to emerge, future research works should consider the interests of servers and clients. The field of social norm formation \cite{Savarimuthu-et-al:2011} can be leveraged to study how clients can form collective opinions about servers (i.e., the task requesters) so that they can persuade them to provide equitable treatment. This development will motivate FL servers to improve their perceived reputations \cite{Yu-et-al:2013} among clients.

\subsection{Trust building through explainability}
Enhancing explainability is useful for dealing with biases in machine learning \cite{fairinDL,Zhang-Yu:2022IJCS}. In FL, explainability research can enhance fairness. From FL clients' perspectives, they lack mechanisms to determine whether they are being treated fairly. Such uncertainty might hamper some clients' future decisions on joining FL. The goal of building explainability is to provide a global understanding on how the FL server makes decisions and how they impact each client's interest in order to build trust between the two parties. However, research in FL explainability must be framed within the context of privacy preservation to avoid conflicting with the primary goal of FL \cite{shokriPrivacyRisksModel2021}.


\section{Conclusions}
In this paper, we provided a comprehensive review of FAFL approaches. We summarized the common assumptions and main notions of fairness adopted in existing FL approaches. We proposed a new taxonomy of FAFL based on the major steps involved in FL, and summarized the challenges faced in each step. After a review of the current studies, we discussed the main evaluation metrics adopted to experimentally measure the performance of FAFL algorithms in order to support long-term sustainability of this field. Finally, we suggest some promising future research directions that can help enhance fairness of future FL approaches. For an interdisciplinary field such as FAFL, collaboration among researchers and industry practitioners from various fields is essential for progress. We hope that this first-of-its-kind survey on FAFL will serve as a useful roadmap towards building FAFL systems.


%



\section*{Acknowledgment}
This research is supported, in part, by the National Research Foundation Singapore and DSO National Laboratories under the AI Singapore Programme (AISG Award No: AISG2-RP-2020-019); Alibaba Group through Alibaba Innovative Research (AIR) Program and Alibaba-NTU Singapore Joint Research Institute (JRI) (Alibaba-NTU-AIR2019B1), Nanyang Technological University, Singapore; the Nanyang Assistant Professorship (NAP); the RIE 2020 Advanced Manufacturing and Engineering (AME) Programmatic Fund (No. A20G8b0102), Singapore; and Future Communications Research \& Development Programme (FCP-NTU-RG-2021-014). Any opinions, findings and conclusions or recommendations expressed in this material are those of the authors and do not reflect the views of the funding agencies.

\ifCLASSOPTIONcaptionsoff
  \newpage
\fi



\bibliography{fafl_bib.bib}
\bibliographystyle{IEEEtran}



%

\vspace{-3em}
\begin{IEEEbiography}[{\includegraphics[width=1in,clip,keepaspectratio]{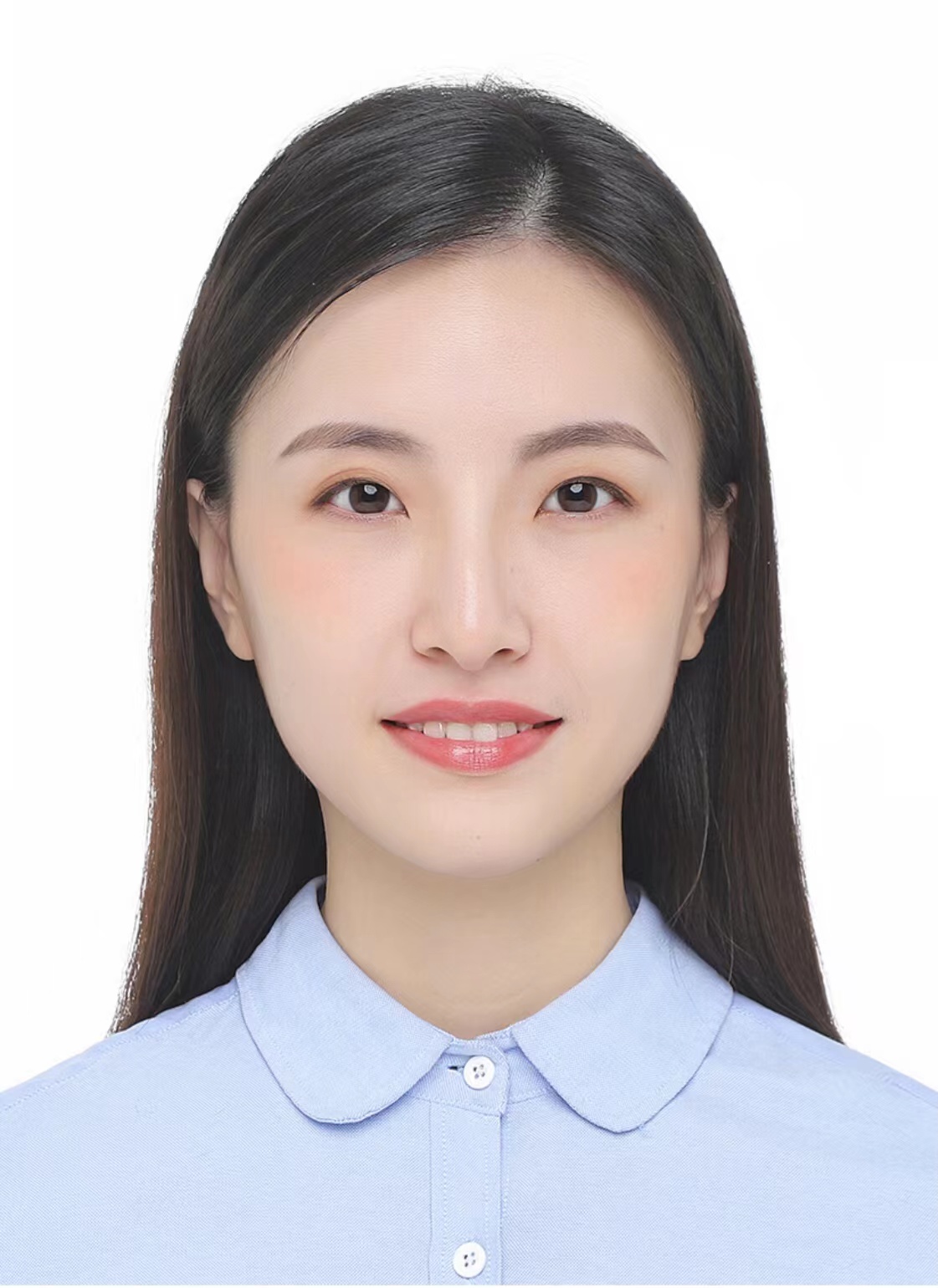}}]{Yuxin Shi} is a PhD scholar at the Alibaba-NTU Singapore Joint Research Institute, Nanyang Technological University (NTU), Singapore. She obtained her Bachelor's degree with Honours (Highest Distinction) in Computer Science from NTU in 2020. Her research focuses on federated learning and machine learning. 
\end{IEEEbiography}
\vspace{-3em}

\begin{IEEEbiography}[{\includegraphics[width=1in,clip,keepaspectratio]{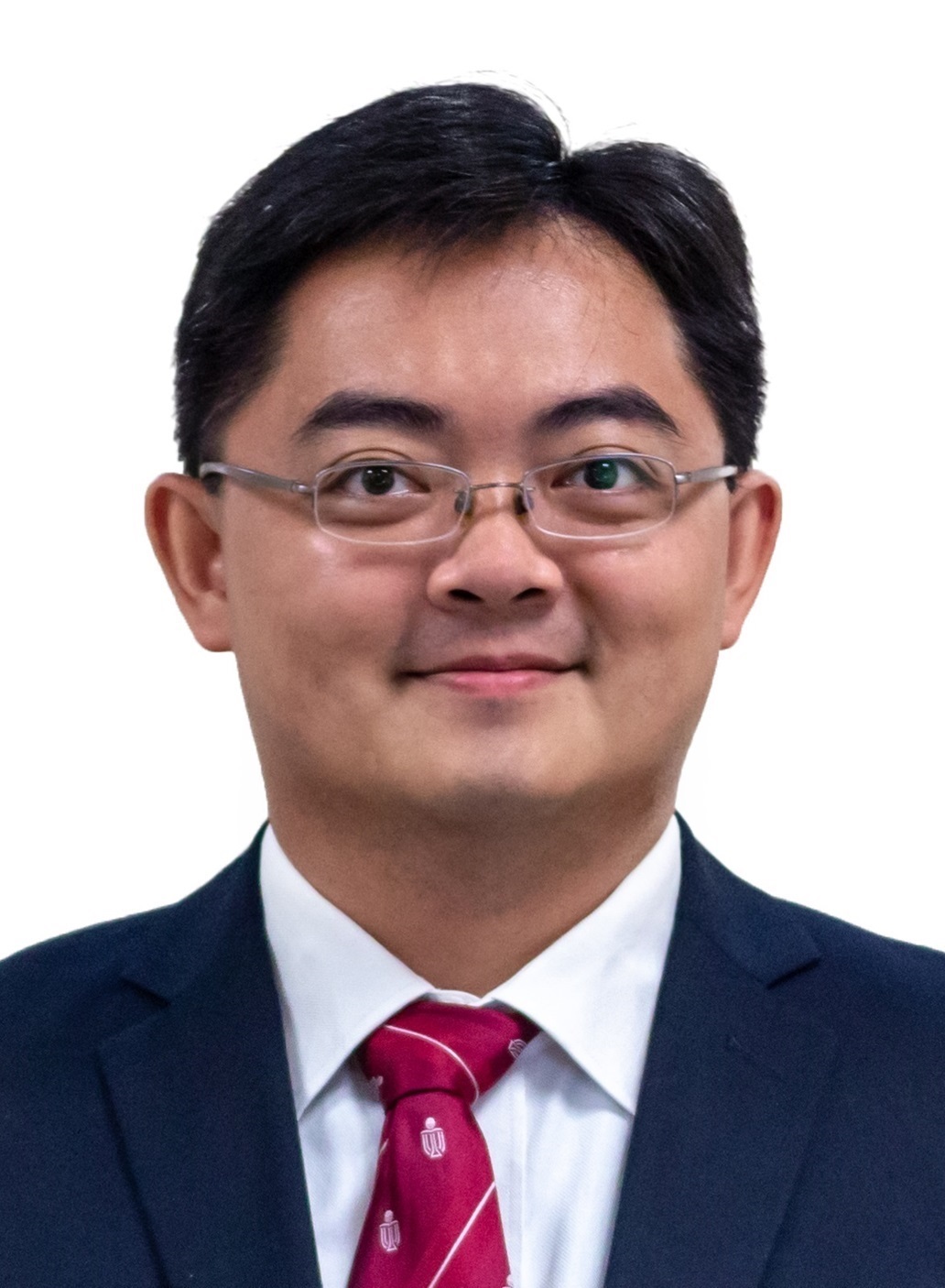}}]{Han Yu} is a Nanyang Assistant Professor (NAP) in the School of Computer Science and Engineering (SCSE), Nanyang Technological University (NTU), Singapore. He held the prestigious Lee Kuan Yew Post-Doctoral Fellowship (LKY PDF) from 2015 to 2018. He obtained his PhD from the School of Computer Science and Engineering, NTU. His research focuses on federated learning and algorithmic fairness. He has published over 200 research papers and book chapters in leading international conferences and journals. He is a co-author of the book \textit{Federated Learning} - the first monograph on the topic of federated learning. His research works have won multiple awards from conferences and journals. He is a \textit{Senior Member} of AAAI, CCF and IEEE.
\end{IEEEbiography}
\vspace{-3em}

\begin{IEEEbiography}[{\includegraphics[width=1in,clip,keepaspectratio]{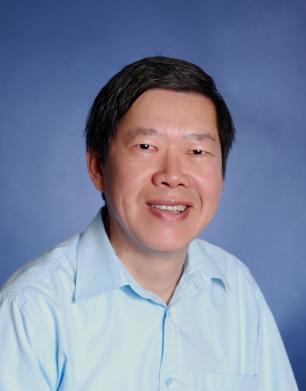}}]{Cyril Leung} is a Professor at the Department of Electrical and Computer Engineering, the University of British Columbia (UBC), Vancouver, Canada. He is a Fellow of the Engineering Institute of Canada. He received his B.Sc.(first class honours) degree from Imperial College, University of London, England, and his M.S. and Ph.D. degrees in electrical engineering from Stanford University. His research interest includes digital communications, wireless communication networks, sensor networks, ubiquitous computing, elderly-friendly technologies, digital/social signal processing, security and privacy, trust computational models, and information theory.

\end{IEEEbiography}




\end{document}